\documentclass[letterpaper, 10 pt, conference]{ieeeconf}  %

\usepackage{graphicx} %
\usepackage{xcolor} %
\usepackage{multirow}
\usepackage{tabularray}
\usepackage{cite}
\usepackage{array}
\usepackage{balance}

\IEEEoverridecommandlockouts
\usepackage{geometry}
\begin{document}

\title{CougarTail \& CUB: A General-Purpose Mast and \\Central Utility Board for Cylindrical Underwater Enclosures}

\author{Ben Washburn$^{*}$, Clayton Smith$^{*}$, Eli Gaskin$^{*}$, Brighton Anderson$^{*}$, \\ Braden Meyers, Brady Moon, Joshua Mangelson
\thanks{*: Equal Contributions}
  \thanks{ This work was funded by the Naval Sea Systems Command (NAVSEA), Naval Surface Warfare Center - Panama City Division (NSWC-PCD) under the Naval Engineering Education Consortium (NEEC) Grant Program under award number N00174-23-1-0005. We additionally thank Tony Zhang for his assistance with vehicle testing and data collection.} 
  \thanks{B. Washburn, C. Smith, B. Anderson, E. Gaskin, B. Meyers, B. Moon, and J.G. Mangelson are at Brigham Young University. They can be reached at \texttt{\{ben1457, cas314, bja1701, egaskin2, bjm255, brady.moon, mangelson\}@byu.edu}.}
}

\maketitle

\begin{abstract}
Cylindrical watertight enclosures are widely used across various underwater systems, from unmanned underwater vehicles (UUVs), to remotely operated vehicles (ROVs), to various sensor platforms. However, electronics are typically built on rectangular PCBs arranged in horizontal stacks, which inefficiently occupy the circular cross-section volume that is critical for both payload capacity and buoyancy management. This paper presents CUB (Central Utility Board) and CougarTail, a general-purpose system designed to address this gap. CUB is a circular PCB sized for 4-inch-diameter enclosures that consolidates a Raspberry Pi Compute Module 5 (CM5) and an STM32 microcontroller, while also providing power management features and auxiliary connections. 
Mounted coaxially, CUB reduces the electronics stack of our CougUV from 200\,mm of tube length and 709\,g to 25\,mm and 
156\,g, returning that length and mass budget to payload and buoyancy trim.
CougarTail is an open-source companion sensor mast that houses a GPS antenna and two dual-band (2.4 and 5 GHz) omnidirectional PCB antennas. Both components are validated through bench testing and integration on a CougUV platform, our small open-sourced torpedo UUVs.
\end{abstract}

\section{Introduction}
\label{sec:intro}

Working with Unmanned Underwater Vehicles (UUVs) or other underwater equipment that contain electronics is difficult. Designing waterproof enclosures is often constrained by limited space and requires time-consuming, expensive custom work. Electronics rapidly consume the available volume inside underwater vehicles, placing tight restrictions on what can be accomplished. Payloads are limited not just by space, but by the need for reliable compute and integration with common sensors and control surfaces — problems that are often solved from scratch for each new vehicle. Electromagnetic waves are also absorbed or disturbed by even a few inches of water, leading to unreliable GPS signals and poor wireless connectivity just below the surface.

In our previous work, we introduced the Configurable Underwater Group of Autonomous Robots (CoUGARs) \cite{COUGARs} as a low-cost, configurable UUV platform for multi-agent autonomy research. The platform consists of several UUVs referred to as CougUVs, open-source vehicles built from 3D-printed and commercial off-the-shelf (COTS) parts with a base cost of \$3,000. As an extension of that work, we present a general-purpose electronics and communications solution aimed at lowering the barrier to entry for underwater robotics and sensing more broadly. By consolidating compute, power management, and connectivity into a compact, low-cost, small form factor package, it is designed to be versatile enough to support a wide range of underwater platforms and research goals, from basic sensing tasks to more complex autonomy applications, giving researchers a working foundation to build on rather than design from scratch.

\begin{figure}[t]
    \centering
    \includegraphics[width=\columnwidth]{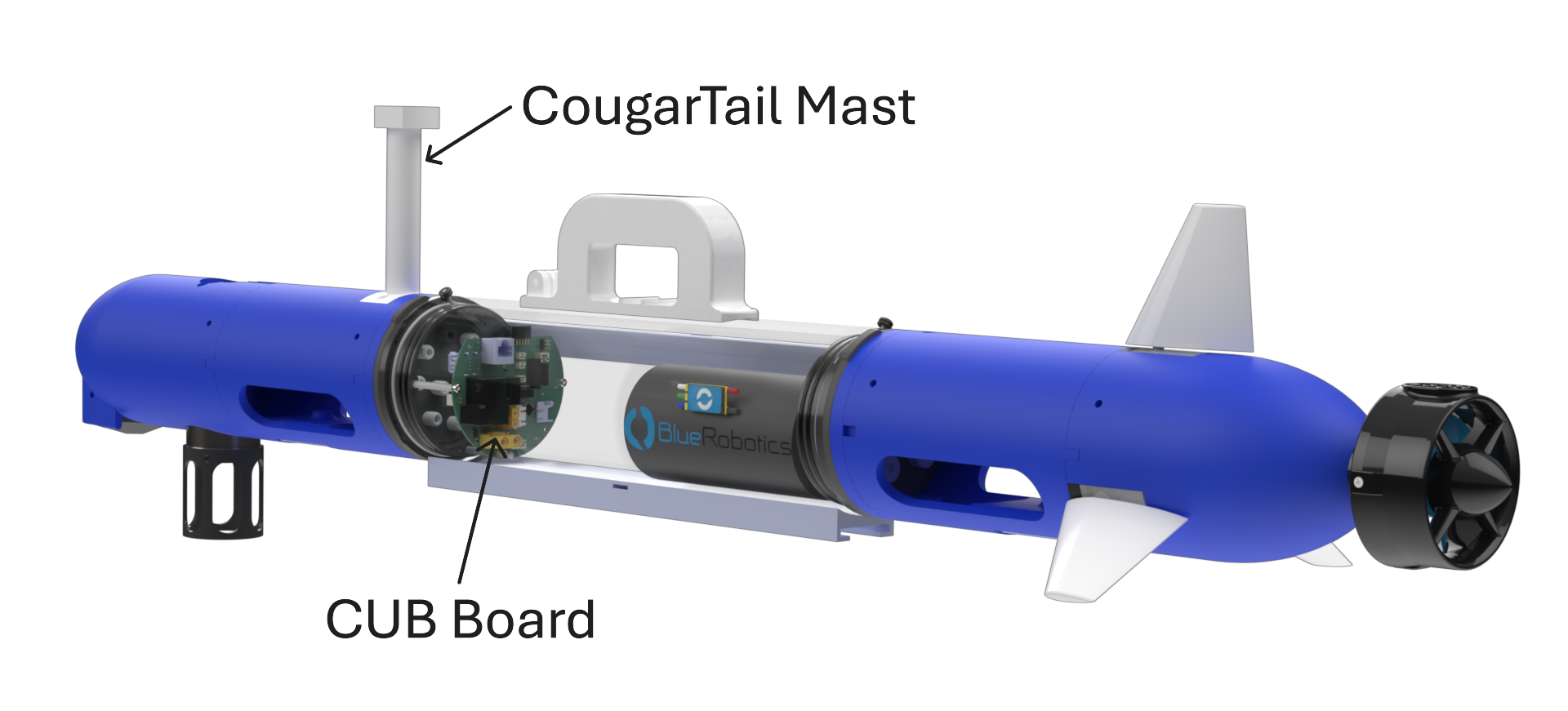}
    \caption{A 3D render of CougarTail and CUB integrated on a CougUV.}
    \label{fig:assembly}
\end{figure}

To provide these tools to the research community, we present the CUB and CougarTail. The CUB is a circular PCB sized for 4-inch-diameter enclosures that consolidates a Raspberry Pi Compute Module 5 (CM5) and an STM32 microcontroller, while also providing power management features and auxiliary connections. CougarTail is an open-source companion sensor mast that houses a GPS antenna and two dual-band (2.4 and 5 GHz) omnidirectional PCB antennas. These components integrate seamlessly with the CougUVs and are also designed to be versatile enough to integrate with many different underwater systems, helping address the challenges of space, compute integration, and communication. Fig.~\ref{fig:assembly} shows both integrated on a CougUV.

\begin{figure*}[t]
    \centering
    \includegraphics[width=\linewidth]{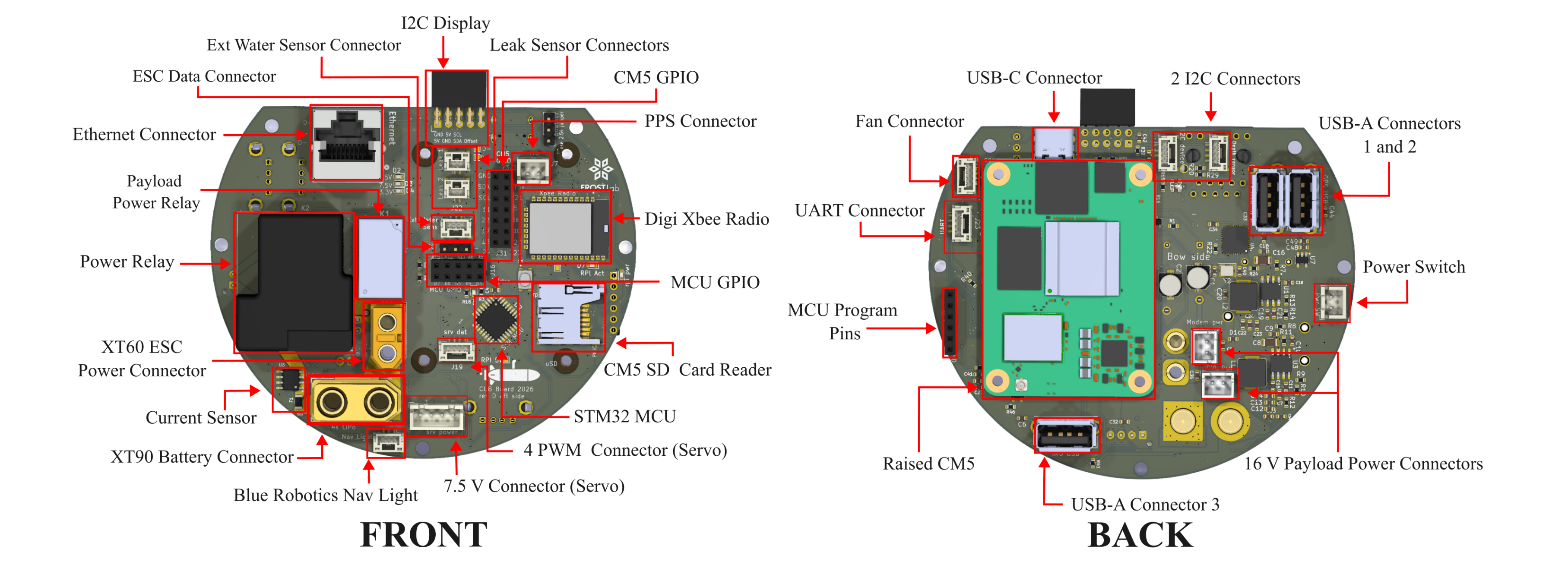}
    \caption{Custom compute board based on the Raspberry Pi Compute Module 5 and STM32,
    designed to fit within a 4-inch Blue Robotics enclosure.}
    \label{fig:pcb}
\end{figure*}

The contributions of this paper include:
\begin{itemize}
    \item A compact, modular compute and I/O board based on the Raspberry Pi Compute Module 5
    \item A custom integrated sensor mast for reliable surface GPS and wireless communications
    \item Validation and testing of both components on the CoUGARs platform
\end{itemize}

\section{Related Work}
\label{sec:related_work}

Low-cost UUVs and ROVs require a suite of electronic components to manage power, sensors, and processing necessary for complex autonomy \cite{bogrekci}. Ensuring electronic components stay dry when submerged is a common challenge for UUVs and ROVs. Cylindrical electronics enclosures, chosen to balance pressure-vessel geometry and container form-factor, are a popular solution \cite{bogrekci}. In addition, open space in the enclosure helps make the UUV buoyant. For low-cost UUVs volume, weight, and accessibility of internal electronics are important considerations.

Most low-cost UUVs use COTS electronic components. Usually electronic components are stacked parallel to the tubes longitudinal axis to accommodate rectangular COTS boards, or generally for simplicity. Designs employing rectangular PCBs include the BlueROV, OpenROV, MIT Spurdog \cite{turrisi2024spurdog}, OASYS \cite{Al-Tawil2023Electronics}, LoCO AUV \cite{loco}, and others \cite{bogrekci,OUbot,Zhou2024TowardsMA,Hu2022LowCost,toolkitUFR}. This choice is not without its tradeoffs. Rectangular boards inefficiently occupy the cross sectional area of the enclosure, and standalone COTS boards with connectors and supports displace a greater amount of space in the enclosure, decreasing ability to adjust buoyancy.
 
Low-cost UUVs or ROVs also employ circular PCBs mounted coaxially with the tube's axis \cite{Al-Tawil2023Electronics, Salem2023Design,mayberry,dubyoski2022DrewUV}. Circular boards occupy less cross-sectional area in cylindrical enclosures, creating space for other components. But the irregular form-factor matching the enclosure's diameter can require a custom design. To aid in low-cost UUV development, we provide a circular mainboard designed for a standard 4-inch enclosure consolidating compute and power management in a single board.

UUVs also require surface communications and GNSS. The ideal location for the necessary antenna is in a mast. A number of mast designs have been developed for low-cost UUVs/ROVs \cite{dubyoski2022DrewUV, trident}. We provide an additional mast as part of the CoUGARs system, prioritizing ease-of-assembly for rapid low-cost UUV prototyping.

\section{Custom PCB Design}
\label{sec:pcb}

The CUB mainboard was designed to consolidate bulky and wiring-intensive power and data systems onto a single easy-to-use board. A cut-circular geometry was chosen to minimize the size footprint and pass-through inconvenience while maximizing board area. The CUB mounts coaxially within commonly-used 4-inch-diameter Blue Robotics acrylic tubes. By integrating previously separate systems onto a single board, CUB achieves a significant reduction in footprint and wiring complexity which increases reliability.
Below is a brief description of the CUB's features.

\subsection{Power Management}
CUB accepts a standard 4S lithium-ion battery through a secure XT90 connection. The board provides a 50 amp fuse and a current sensor for safety and systems monitoring. The board can be easily turned on and off from outside its enclosure through a main power switch broken out to a 2-pin JST connector. Servos are powered through a 4-pin JST header, providing 7.5V at up to 5 Amps. A USB-C connection is also included to power and program the CM5 without powering the entire board.
 
\subsection{Embedded Processing System}
The mainboard embeds a fully Linux-capable CM5, enabling processing-power hungry autonomous systems to run continuously without external connection. The CM5 sits on a raised mount, allowing for easy replacement in case of device failure. To control processor temperature, a standard CM5 heatsink may be installed. For ease-of-use, files are stored on an easily replaceable SD card. The CM5 communicates with a STM32 microcontroller over UART. The STM32 allows for expanded connectivity and offloads lower level tasks. Communication with various sensors over General Purpose Input Output pins (GPIOs) and servo control via Pulse Width Modulation (PWM) pins are handled by the microcontroller. Both processors can be setup with CoUGARs software.

\subsection{Peripherals}
The board supports a pair of leak sensors, an external ESC, 4 fin-control servos, an exterior water sensor, and an attachable status display. A Digi XBee radio allows for long range communication and is compatible with the CougarTail mast. The CougarTail 2.4 GHz antennas easily attach to the CM5 and the radio through a UFL connection. Extra I2C, UART, Ethernet, and USB-A connections are also provided for experimentation purposes.

\subsection{Development Support Features}
An onboard relay controls two power sockets each capable of supporting a payload up to 3 Amps. Payload power control coupled with the exterior water sensor provides the capability to manage acoustic sensors that overheat when run in air. An easily-accessible header provides 3.3V and 5V power, 10 STM32 GPIOs, and 9 CM5 GPIOs. STM32 and CM5 connections allow for I2C, SPI, UART and CAN communication. Built-in indicator LEDs, debug LEDs and programming headers can be used to quickly diagnose and fix board-specific issues in testing and in the field.

The variety of features embedded in CUB in a convenient form factor make it easy to both prototype with and implement in a completed system. Sensing payloads, different robotic systems, and communication systems can be added with ease.

\begin{figure}[t]
    \centering
    \includegraphics[width=0.95\columnwidth]{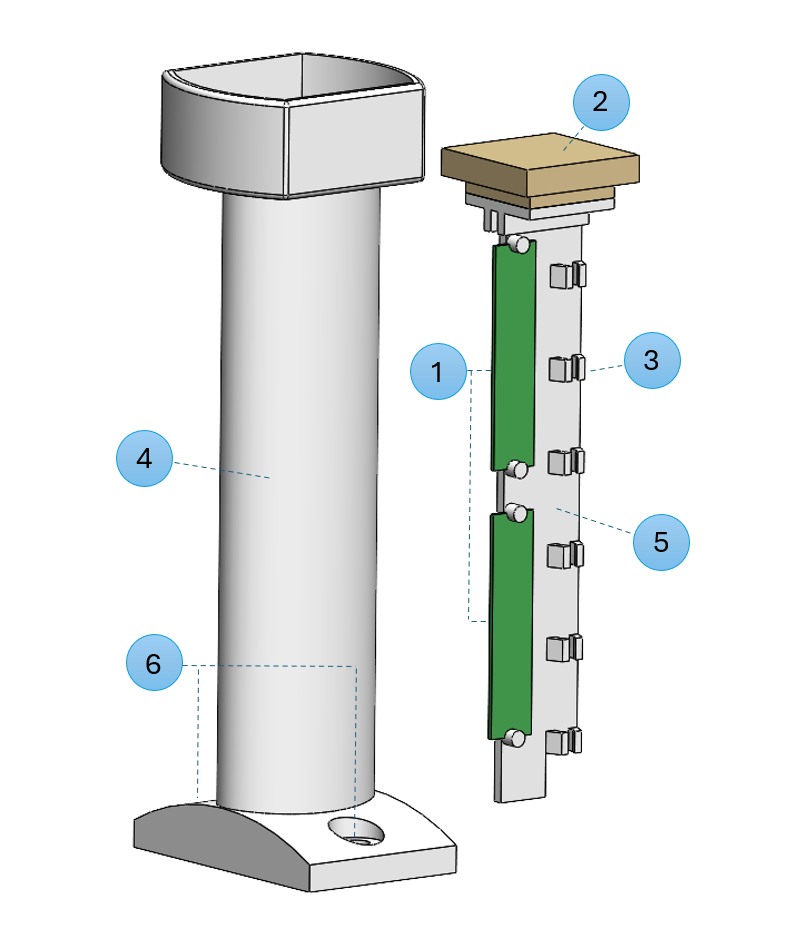}
    \caption{Molex 146187 dual-band antennas (1), Taoglas AP.25F.07.0078A GNSS antenna (2), Wire clips (3), Mast housing (4), Mast insert (5), Mounting screw holes (6). Labeled render of CougarTail mast housing (left) and insert (right). The mast insert was placed in the housing and filled with epoxy. Cables were secured by clips and fed through a polyurethane tube through the base of the mast housing (not shown).}
    \label{fig:mast_exploded}
\end{figure}

\section{Integrated Sensor Mast}
\label{sec:mast}

Surface operations with our CougUV platform \cite{COUGARs} previously relied on GPS and radio antennas housed within the main enclosure. Even shallow submersion attenuated these signals enough to render them unreliable, forcing the vehicle to fully surface and orient itself favorably before acquiring a GPS fix or establishing a communication link. To address this, we designed a custom 100\,mm-tall mast that extends above the waterline whenever the vehicle is surfaced (see Fig.~\ref{fig:mast_exploded}). The housing was kept as compact as possible while still accommodating three antennas to avoid unnecessary drag. It carries one ceramic GPS antenna and two Molex PCB antennas, one each for WiFi and RF communication.

A central design challenge was achieving a reliable, long-term watertight seal around the antennas without degrading their RF performance. The three antennas were first attached to a thin plastic mounting plate that fixed their relative positions and orientations, then dip-coated in epoxy. This dip-coating process fully encapsulated the sensitive electronics and solder joints in a continuous, bubble-free layer while preserving the precise antenna orientation established on the plate -- critical for GPS reception and radio pattern performance. The coated assembly was then inserted into a mast shell, FDM 3D printed from PETG at 100\% infill for maximum strength and minimal porosity, and the remaining internal volume was backfilled with epoxy resin to add rigidity and provide a secondary barrier against water. The epoxy coating provides a theoretical depth rating of more than 100 meters. The GPS antenna sits at the top of the stack, giving it an unobstructed view of the sky, while the WiFi and RF antennas are positioned lower in the housing. All three antennas connect via coaxial cable that exits the mast through a polyurethane tube and enters the main enclosure through a sealed cable penetrator, keeping the electrical connections outside the mast itself dry and serviceable.

The mast is held in place by a custom fore extension piece, also FDM-printed from PETG, that interfaces mechanically between the nosecone and the central enclosure. When surfaced, operators communicate with the vehicle primarily over RF radio, which offers sufficient range and throughput for transmitting mission parameters and startup signals. WiFi is reserved for direct SSH access when the operator is near the vehicle and finer-grained configuration or debugging is required. Underwater communication continues to rely on the SeaTrac X150 USBL acoustic modem, as described in our previous work \cite{COUGARs}.

\section{Testing and Results}
\label{sec:results}

The CUB and CougarTail were thoroughly bench tested, with deployment onto the CougUV platform in progress. The results below detail the performance of both systems and the improvements they provide to the CougUV platform.

\subsection{Space and Weight Improvements}
The electronics system CUB replaces occupied 200\,mm of longitudinal 
space within the body tube and weighed 709\,g. CUB occupies 25\,mm 
and weighs 156\,g, reductions of 87.5\% and 78\% respectively. 
The 175\,mm of tube length recovered is available for sensor 
payloads, ballast, or buoyancy foam, and the 553\,g removed relaxes
the trim budget.

\begin{table}[t]
    \centering
    \renewcommand{\arraystretch}{1.3}
    \begin{tabular}{| >{\centering\arraybackslash}m{2.2cm} | >{\centering\arraybackslash}m{4.8cm} |}
        \hline
        \textbf{Metric} & \textbf{Result} \\
        \hline
        Horiz. accuracy (1-sigma) & 1.07 m mean / 0.57 m median \\
        \hline
        Stability (rolling std of accuracy) & 0.044 m \\
        \hline
        Fix reliability & 100\% (0 dropouts, 232.6 s @ 5.0 Hz) \\
        \hline
    \end{tabular}
    \caption{GNSS accuracy summary.}
    \label{tab:gnss-accuracy}
    \vspace{-2em}
\end{table}

\begin{figure}[t]
    \centering
    \includegraphics[width=0.9\columnwidth]{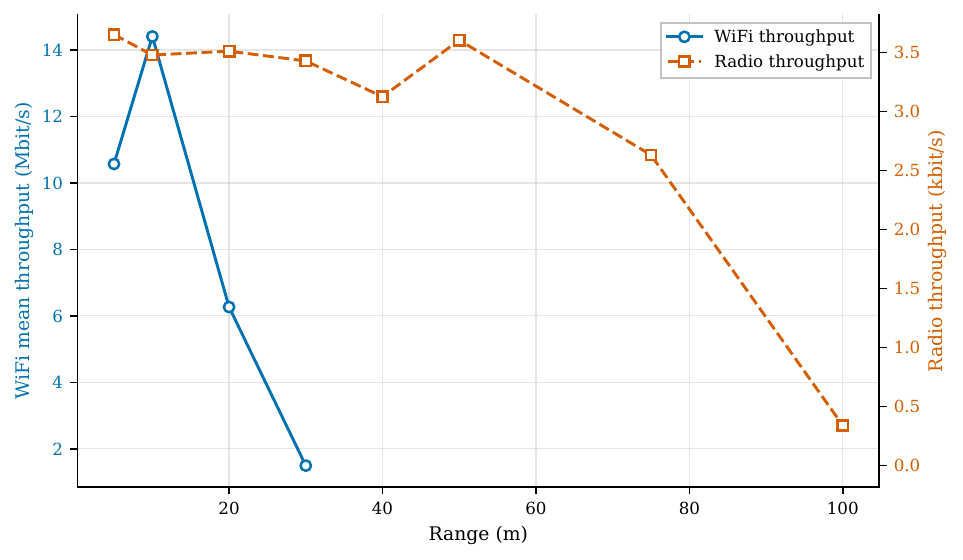}

    \caption{CougarTail performance summary. GNSS accuracy was recorded continuously over the course of a representative mission, while WiFi and radio throughput were measured at controlled standoff distances ranging from 5\,m to 100\,m.}
    \label{fig:cougartail_summary}
\end{figure}

\begin{figure}[t]
    \centering
    \includegraphics[width=0.8\columnwidth]{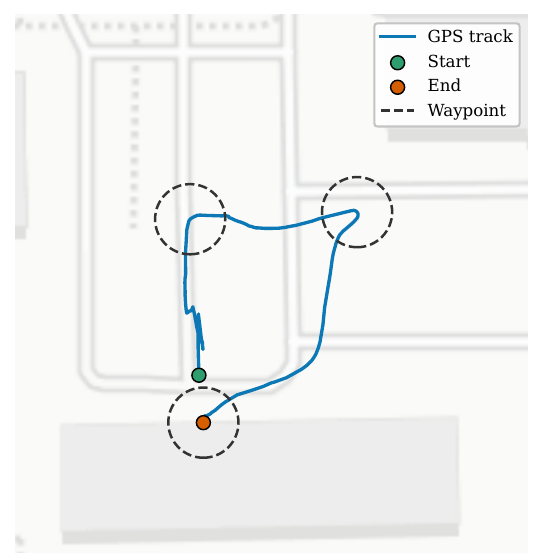}
    \caption{Cart-based waypoint-following test of the CUB and CougarTail. Positions were recorded while the CUB processed live sensor input and actuator commands, confirming end-to-end functionality of the sensing, control, and communication pipeline.}
    \label{fig:CUB_mission_plot}
\end{figure}

\subsection{Mast Data Rates and GPS Performance}
The mast's water-tightness was validated by submerging the 
assembly. Following submersion, mast data rates measured above water showed no degradation relative to pre-submersion baseline, confirming the epoxy-and-shell seal prevented water ingress to the antennas and internal wiring. In future work, the mast will be pressure tested to greater depths to find its actual depth rating.

The throughput of the CougarTail was tested at variable distances between 5 and 100 meters. These tests were carried out mounted on a CougUV interfacing with the CUB. Field communication range tests were conducted by recording a GPS location for both the CougUV and a stationary base station computer at each of several specified distances. Two independent link types were evaluated in parallel: a WiFi link between the CM5's onboard PCB antenna and the base station, and a radio link between a Digi XBee 3 module on the CUB and a companion Digi XBee module at the base station. For both of these links, throughput was measured at each point. The resulting throughput as a function of range is presented in Fig.~\ref{fig:cougartail_summary}. 

The GPS antenna in the CougarTail was evaluated while connected to a SBG Systems Ellipse-N INS on a cart, in trajectories similar to missions that would take place in the operation of the CougUV. Table~\ref{tab:gnss-accuracy} summarizes the result: over a 232.6\,s 
trial the antenna held a 1.07\,m mean horizontal accuracy at 100\% 
fix reliability, with no dropout gaps observed.

\subsection{Mission Results}
A comprehensive functional test of the CUB and CougarTail was performed by mounting the system on a cart and executing waypoint missions, exercising the full sensor and communication pipeline under realistic timing conditions. This validated the board's ability to ingest sensor data, execute the waypoint-following logic, and generate corresponding vehicle control commands in real time. Fig.~\ref{fig:CUB_mission_plot} shows the commanded waypoint path alongside the recorded vehicle-frame position during one such test.
In-water waypoint missions with the CougUV are planned as future work to validate closed-loop control performance under real hydrodynamic disturbances.

\section{Conclusion}
\label{sec:conclusion}

This paper presents CougarTail and CUB, a circular PCB and integrated sensor mast designed for 4-inch-diameter cylindrical underwater enclosures. By matching the board geometry to the enclosure cross-section and consolidating compute, power management, and peripheral connectivity onto a single board, CUB reduces wasted volume compared to conventional rectangular PCB stack-based systems. CougarTail complements the board with a co-designed mast that provides GPS, radio, and WiFi for reliable surface communications. CougarTail designs and assembly instructions are available at https://frost-lab.gitbook.io/cougars/. To request access to the CUB designs, please contact the authors.

\vfill
\balance
\bibliographystyle{IEEEtran}
\bibliography{ref}

\end{document}